\documentclass[10pt,twocolumn,letterpaper]{article}
\pdfoutput=1
\usepackage{url}                % for URL
\usepackage{times}
\usepackage{epsfig}
\usepackage{graphicx}
\usepackage{amsmath}
\usepackage{amssymb}
\usepackage{subfigure}          % for sub-figure
\usepackage{multirow}           % for complex table
\usepackage{bm}

\usepackage[pagebackref=true,breaklinks=true,letterpaper=true,citecolor=blue,colorlinks,bookmarks=false]{hyperref}

\begin{document}

%%%%%%%%% TITLE
\title{Anisotropic Diffusion-based Kernel Matrix Model \\for Face Liveness Detection}

\author{Changyong Yu, Yunde Jia\\
Beijing Laboratory of Intelligent Information Technology, \\School of Computer Science,
Beijing Institute of Technology, \\Beijing 100081, P.R. China\\
{ Email:\{yuchangyong,jiayunde\}@bit.edu.cn}
% For a paper whose authors are all at the same institution,
% omit the following lines up until the closing ``}''.
% Additional authors and addresses can be added with ``\and'',
% just like the second author.
% To save space, use either the email address or home page, not both
}
\date{}
\maketitle
%\thispagestyle{empty}

%%%%%%%%% ABSTRACT
\begin{abstract}
Facial recognition and verification is a widely used biometric technology in security system. Unfortunately, face biometrics is vulnerable to spoofing attacks using photographs or videos. In this paper, we present an anisotropic diffusion-based kernel matrix model (ADKMM) for face liveness detection to prevent face spoofing attacks. We use the anisotropic diffusion to enhance the edges and boundary locations of a face image, and the kernel matrix model to extract face image features which we call the diffusion-kernel (D-K) features. The D-K features reflect the inner correlation of the face image sequence. We introduce convolution neural networks to extract the deep features, and then, employ a generalized multiple kernel learning method to fuse the D-K features and the deep features to achieve better performance. Our experimental evaluation on the two publicly available datasets shows that the proposed method outperforms the state-of-art face liveness detection methods.
\end{abstract}

%%%%%%%%% BODY TEXT
\section{Introduction}

Face recognition and verification \cite{face1, face2} has become the most popular technology in high-level security systems due to the natural, intuitive, and less human-invasive face biometrics. Unfortunately, face biometrics is vulnerable to spoofing attacks using photographs or videos of the actual user. Attackers can attempt to hack the security system by using printed photos, mimic masks, or screenshots, and they can also use the captured or downloaded video sequences containing facial gestures like eye blinking of the valid user to invade the security system. In order to mitigate this problem, many researchers have made much effort to face liveness detection based on image quality \cite{sta3, sta2}, spectrum \cite{spectrum1, spectrum2}, and motion information like eye blinking \cite{eyeblink}, mouth movement \cite{mousemovement}, and head pose \cite{headpose}. Recently, Diffusion methods have been applied to face liveness detection \cite{diffspeed, diffCNN}, which can estimate the difference in surface properties between live and fake faces and achieve spectacular performance. However, it still remains a big challenge to detect face liveness against spoofing attacks.

In this paper, we present an anisotropic diffusion-based kernel matrix model (ADKMM) for face liveness detection to prevent face spoofing attacks. The ADKMM can accurately estimate the difference in surface properties and inner correlation between live and fake face images. Figure 1 illustrates the overview of our method. The anisotropic diffusion is used to enhance the edges and boundary locations of a face image sequence, and the kernel matrix model is used to extract face features from the sequence which reflect the inner correlation of the sequence. We call these features the diffusion-kernel (D-K) features. To achieve better performance against the spoofing attack, we also extract the deep features using deep convolution neural networks, and then utilize a generalized multiple kernel learning method to fuse the D-K features and the deep features. The deep features can work well with D-K features by providing complementary information.
\begin{figure*}[htb]
\begin{minipage}[b]{1.0\linewidth}
  \centering
  \centerline{\includegraphics[width=6.4 in]{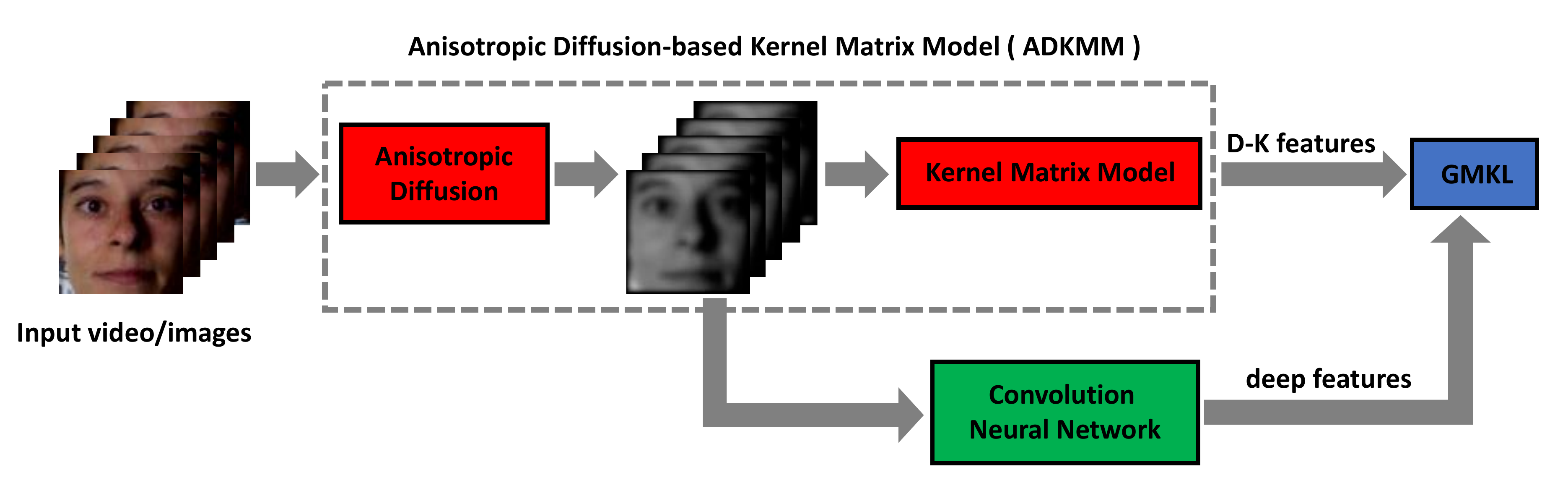}}
\end{minipage}
\caption{An illustration of the proposed method. We diffuse the input video or images by anisotropic diffusion method first, next we use kernel matrix model to extract the diffusion-kernel (D-K) features. Then, we extract deep features from the diffused images by deep convolution neural networks. Finally, the D-K features and deep features are fused by a generalized multiple kernel learning (GMKL) method for face liveness detection to prevent face spoofing attacks.}
\end{figure*}
Synthetical considering these two kinds of feature relationships, we can capture the differences effectively in both sides of illumination characteristics and inner correlation of face images. The experimental results on various publicly available datasets demonstrate that our method provides a reliable performance of face liveness detection.

The main contributions of this paper can be summarized as follows:

$\bullet$ We present an anisotropic diffusion-based kernel matrix model (ADKMM) that can accurately estimate the difference in surface properties and inner correlation between live and fake face images to extract diffusion-kernel (D-K) features for face liveness detection.

$\bullet$ We utilize a generalized multiple kernel learning method to fuse the D-K features and the deep features extracted by deep convolution neural networks to get better performance against the spoofing attack.

$\bullet$ Our method achieves an impressive accuracy on the publicly available datasets and outperforms the state-of-art face liveness detection methods.

\section{Related Work}

Many methods of face liveness detection are based on the analysis of a single image. These methods assume that fake faces tend to lose more information by the imaging system and thus come into a lower quality image under the same capturing condition. Li et al. \cite{sta1} proposed to analyze the coefficients of Fourier transform since the reflections of light on 2D and 3D surfaces result in different frequency distributions. For example, fake faces are mostly captured twice by the camera, so their high-frequency components are different with those of real faces. Zhang et al. \cite{sta2} attempted to extract frequency information using multiple DoG filters to detect the liveness of the captured face image. Tan et al. \cite{sta3} and Peixoto et al. \cite{sta4} combined the DoG filter and other models to extract efficient features from the input image to improve the liveness detection performance. Maatta et al. \cite{sta5} extracted the micro texture of input image using the multi-scale local binary pattern (LBP). Based on these micro textures, they used SVM classifier to detect the face liveness. Kim et al. \cite{diffspeed} calculated the diffusion speed of a single image, then they used a local speed model to extract features and input them into a linear SVM classifier to distinguish the fake faces from real ones. Alotaibi et al. \cite{diffCNN} used nonlinear diffusion to detect edges in the input image and utilized the diffused image to detect face liveness by using convolution neural networks.

\begin{figure}
\begin{minipage}[b]{1.0\linewidth}
  \centering
  \centerline{\includegraphics[width=3 in]{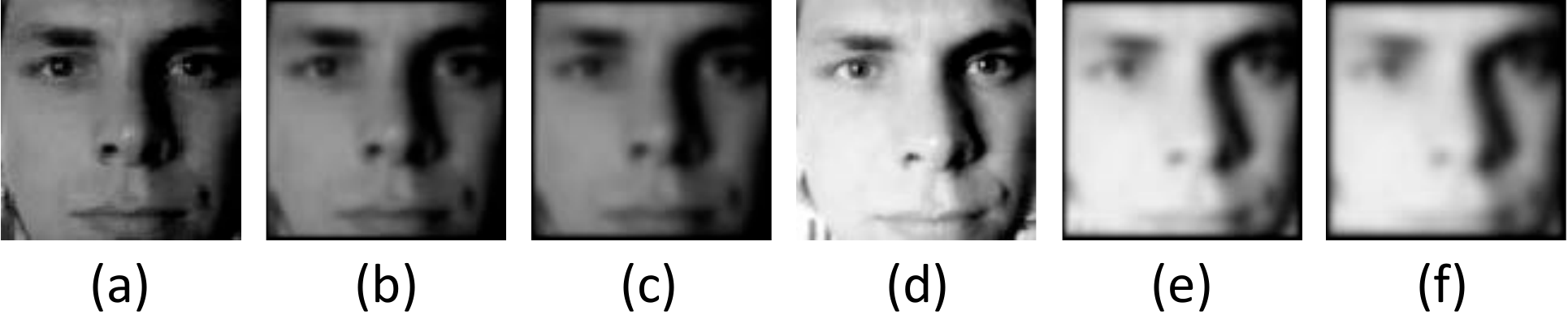}}
\end{minipage}
\caption{Examples of diffused images with different iteration numbers $(L)$. (a), (b) and (c) are live face images, (d), (e) and (f) are fake face images. (a) and (d) Original images. (b) and (e) Diffused image with $L = 5$. (c) and (f) Diffused image with $L = 15$.}
\end{figure}

Motion-based approaches are another common class of face liveness detection, which aim at detecting subconscious response of a face. Given an image sequence, these methods attempt to capture facial response like eye blinking, mouth movement, and head pose, then exploit spatial and temporal features. Pan et al. \cite{eyeblink} detected the eye blinking behavior using a unidirectional conditional graphic framework. A discriminative measure of eye states is also incorporated into the framework to detect face liveness. Singh et al. \cite{dyn2} applied the haar classifier and distinguished the fake faces from the real ones by detecting eye and mouth movements. Anjos et al. \cite{dyn3} attempted to detect motion correlations between the user’s head and the background regions obtained from the optical flow that indicate a spoofing attack. Tirunagari et al. \cite{dyn4} used the dynamic mode decomposition (DMD) algorithm to capture face from the input video and extract dynamic visual information to detect the spoofing attack. Bharadwaj et al. \cite{dyn5} proposed a novel method to detect face liveness using the configuration LBP and motion estimaion to extract the facial features. Since multiple input frames are required to track face, dynamic approaches usually cost more detection time and computing resources. Besides, some dynamic methods require user to follow some instructions which in the users’ experience is inconvenient.

Different from existing methods, our work focus on the the difference in surface properties and inner correlation between live and fake face images. We present an anisotropic diffusion-based kernel matrix model (ADKMM) for face liveness detection to prevent face spoofing attacks.  The anisotropic diffusion method helps to distinguish fake faces from real ones by diffusing the input image to enhance the difference in illumination characteristics. From these images, we can gain the depth information and boundary locations of face images. The D-K features extracted by kernel matrix model can significantly reflect the inner correlation of the faces image sequences and thus lead to a better classification.

\section{Proposed Method}

Our method uses the anisotropic diffusion to enhance the edges and boundary locations of a face image, and the kernel matrix model to extract the diffusion-kernel (D-K) features. To achieve better performance against the spoofing attacks, we extract the deep features by deep convolution neural networks, and utilize a generalized multiple kernel learning method to fuse the D-K features and the deep features.

\subsection{Anisotropic Diffusion}
The  anisotropic diffusion method is used to diffuse the input video or images to enhance edge information. Some examples of
diffused images are shown in Figure 2. Consider the anisotropic diffusion equation \cite{Adiff}
$$I_t=div(c(x,y,t) \nabla I)=c(x,y,t) \Delta I + \nabla c\cdot \nabla I, \eqno{(1)}$$
where the $div$ represents the divergence operator, and $\nabla$ , $\Delta$ represent the gradient and Laplacian operators, respectively. $I_0(x,y)$ is the original image and $t$ is the variance in Gaussian kernel $G(x, y, t )$. It reduces to the isotropic heat diffusion equation $I_t=c\Delta I$ if $c(x, y, t)$ is a constant. Suppose that at time $T$, we knew the locations of the region boundaries or edges for that scale. We want to encourage a region to be smooth in preference to smoothing across the boundaries. We could achieve this by setting the conduction coefficient to be 1 in the interior of each region and 0 at the boundaries. The blurring would then take place separately in each region with no interaction between regions and the region boundaries would remain sharp.

The next task is to localize the region boundaries at each scale. Perona et al. \cite{Adiff} compute the best estimate of the location of the boundaries appropriate to that scale. Let $E(x,y,t)$ be such an estimate, the conduction coefficient $c(x,y,t)$ can be chosen to be a function $c = g(||E||)$ of the magnitude of $E$. $E=0$ means the points are in the interior of a region and in other cases means the points are at the edge. Specially, if the function $g ( . )$ is chosen properly, the diffusion in which the conduction coefficient is chosen locally as a function of the magnitude of the gradient of the brightness function, i.e.,
$$c(x,y,t)=g(||\nabla I(x,y,t)||) \eqno{(2)}$$
will not only preserve, but also sharpen the brightness edges.

\subsubsection{Edge Enhancement}
The blurring of edges is the main price paid for eliminating the noise with conventional low-pass filtering and linear diffusion. It will be difficult to detect and localize the region boundaries. Edge enhancement and reconstruction of blurry images can be obtained by high-pass filtering or running the diffusion equation backwards in time. If the conduction coefficient is chosen to be an appropriate function of the image gradient, we can make the anisotropic diffusion enhance edges while running forward in time, thus ensuring the stability of diffusions.

The model make the edge as a step function convolved with a Gaussian kernel. Without loss of generality, assume that the edge is aligned with the $y$ axis.

The expression for the divergence operator simplifies to
$$div(c(x,y,t) \nabla I)=\frac {\partial}{\partial x}(c(x,y,t)I_x). \eqno{(3)}$$
We choose $c$ to be a function of the gradient of $I: c ( x, $ $y, t ) =  g ( I_x ( x, y, t ))$ as in Eq.(2). Let $\phi( I_x ) \doteq g( I_x ) \cdot I_x$ denote the flux $c\cdot I_x$.

Then the 1-D version of the Eq.(1) is
$$I_t=\frac {\partial}{\partial x}\phi (I_x)=\phi ' (I_x) \cdot I_{xx}. \eqno{(4)}$$
The variation of the edge slope is $\partial / {\partial t(I_x)}$. If $ c(\cdot)>0$, the function $ I(\cdot)$ is smooth, and the order of differentiation may be inverted as
\begin{equation}
\begin{split}
\frac {\partial}{\partial t}(I_x) &= \frac {\partial}{\partial x}(I_t)=\frac {\partial}{\partial x}\left (\frac {\partial}{\partial x}\phi (I_x)\right )\\
&= \phi '' \cdot I^2_{xx}+\phi ' \cdot I_{xxx}.
\end{split}
\tag{5}
\end{equation}

Suppose the edge is oriented as $I_x > 0$. At the point of inflection $I_{xx} = 0$, and $I_{xxx} << 0$ since the point of inflection corresponds to the point with maximum slope. Then in a neighborhood of the point of inflection $\partial / \partial t(I_x)$ has sign opposite to $\phi ^{'} ( I_x )$. If $\phi ^{'} ( I_x )>0$ the edge slope will decrease with time, on the contrary $\phi ^{'} ( I_x )<0$ the slope will increase with time. Based on the increase of the slope, the edge becomes sharper.

\subsubsection{Diffusion Scheme}
The anisotropic diffusion and edge detection method \cite{Adiff} we utilized is a simple numerical scheme that is described in this section.

Eq.(1) can be discretized on a square lattice with brightness values associated to the vertices, and conduction coefficients to the arcs. A 4-nearest neighbors discretization of the Laplacian operator can be given by
\begin{equation}
\begin{split}
I^{L+1}_{i,j} &= I^{L}_{i,j} + \lambda[c_N \cdot \bigtriangledown_N I + c_S \cdot \bigtriangledown_S I \\
&+c_E \cdot \bigtriangledown _E I + c_W \cdot \bigtriangledown _W I]^L_{i,j}
\end{split}
\tag{6}
\end{equation}
where $0 \leq \lambda \leq 1/4$ for the numerical scheme to be stable, $N$, $S$, $E$, $W$ are the mnemonic subscripts for North, South, East, West. The symbol $\bigtriangledown $ indicates nearest-neighbor differences:
\begin{equation}
\begin{split}
&\bigtriangledown _{N}I_{i,j}\equiv I_{i-1,j}-I_{i,j}\\
&\bigtriangledown _{S}I_{i,j}\equiv I_{i+1,j}-I_{i,j}\\
&\bigtriangledown _{E}I_{i,j}\equiv I_{i,j+1}-I_{i,j}\\
&\bigtriangledown _{W}I_{i,j}\equiv I_{i,j-1}-I_{i,j}.
\end{split}
\tag{7}
\end{equation}

In some directions, if the difference is obviously, it indicates that this point may at the edge and we should preserve and sharpen the edge information. Figure 3 shows the overall scheme to detect the edges and preserve them.

The value of the gradient can be computed by
\begin{equation}
\begin{split}
&c^L_{N_{i,j}}=g(|\bigtriangledown _{N}I^L_{i,j}|)\\
&c^L_{S_{i,j}}=g(|\bigtriangledown _{S}I^L_{i,j}|)\\
&c^L_{E_{i,j}}=g(|\bigtriangledown _{E}I^L_{i,j}|)\\
&c^L_{W_{i,j}}=g(|\bigtriangledown _{W}I^L_{i,j}|).
\end{split}
\tag{8}
\end{equation}

If the $\bigtriangledown$ in some directions change greater, the value of the $c(\cdot)$ should be smaller, and thus will preserve and sharpen the boundary locations.

In this scheme, the conduction tensor in the diffusion equation is diagonal with entries $g (|I_x|)$ and $g (|I_y|)$ instead of $g (||\nabla I||)$ and $g (||\nabla I||)$. This diffusion scheme preserves the property of the continuous Eq.(1) that the total amount of brightness in the image is preserved.

\subsection{Kernel Matrix}
 Kernel matrix has recently received increasing attention as a generic feature representation in various recognition and classification tasks. For a large set of kernel functions, the kernel matrix is guaranteed to be nonsingular, even if samples are scarce. More importantly, kernel matrix gives us unlimited opportunities to model nonlinear feature relationship in an efficient manner.

 \begin{figure}
\begin{minipage}[b]{1.0\linewidth}
  \centering
  \centerline{\includegraphics[width=3 in]{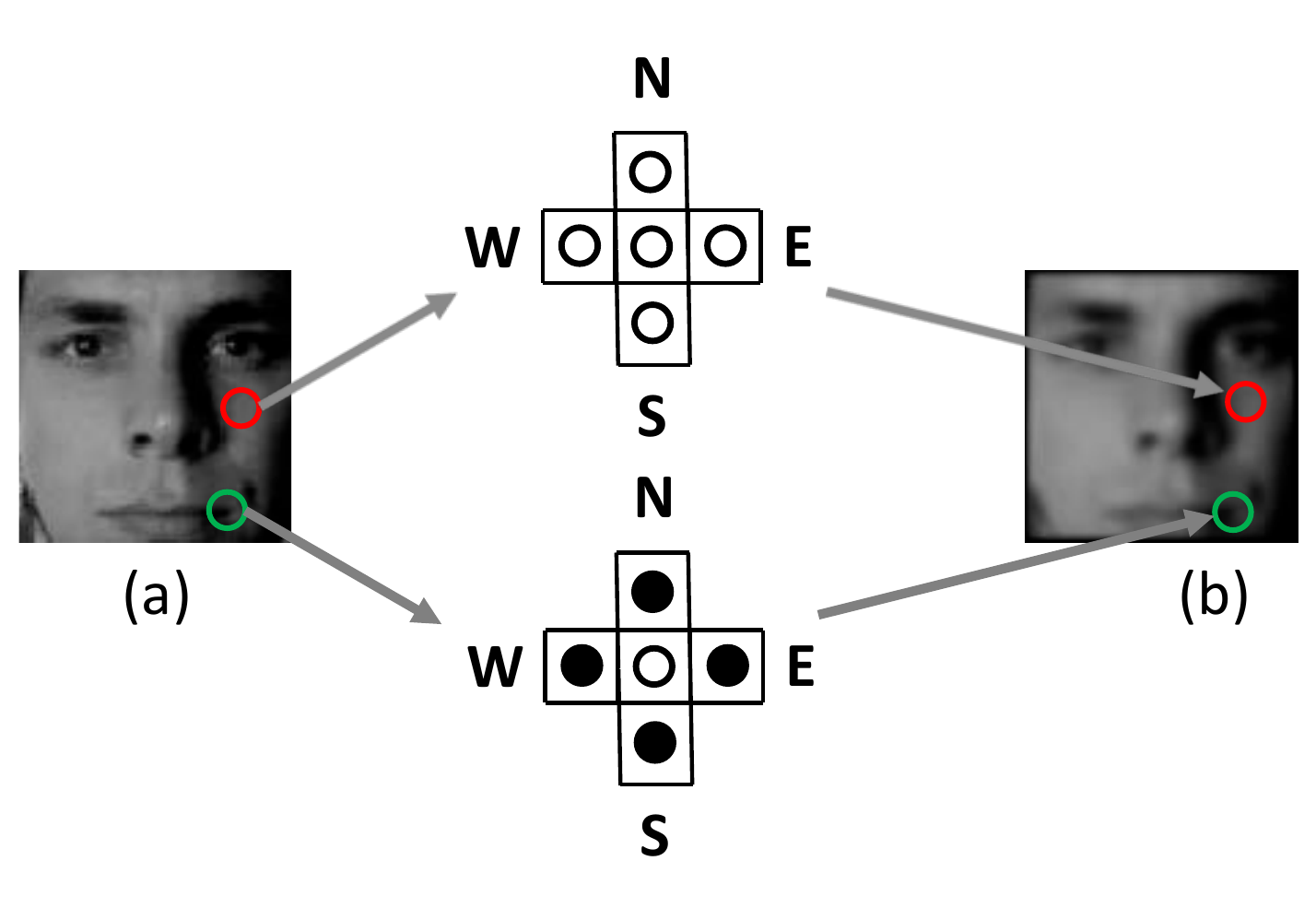}}
\end{minipage}
\caption{(a) Original image. (b) Diffused image. The point in red circle is similar with its nearest-neighbors that indicates this point is in the interior of a region. The point in green circle has distinct differences with its nearest-neighbors that indicates this point is at the edge.}
\end{figure}

\subsubsection{Feature Representation}
We use the kernel matrix, M, as a generic feature representation \cite{KM}. The $(i, j)$ th entry of M is defined as
$$k_{i,j}=<\phi (f_i),\phi (f_j)>=\kappa (f_i,f_j), \eqno{(9)}$$
where $\phi(\cdot)$ is an implicit nonlinear mapping and $\kappa (\cdot , \cdot)$ is the induced kernel function. The mapping $\kappa (\cdot , \cdot)$ is applied to each feature $f_i$, rather than to each sample $x_i$ as usually seen in kernel-based learning methods. The most significant advantage of using M is that we can have much more flexibility to efficiently model the nonlinear relationship among features.

We can evaluate the similarity of feature distributions by some specific kernels like Bhattacharyya kernel \cite{Bhkernel}. When we do not know beforehand what kind of nonlinear relationship shall be modeled, we can apply a kernel representation. In this case, any general purpose kernel, such as the Gaussian RBF kernel
$$\kappa (f_i , f_j)=exp(-\gamma ||f_i-f_j||^2) \eqno{(10)}$$
can be employed. Also, once it becomes necessary, users are free to design new, specific kernels to serve their goals. Using a kernel matrix as feature representation is so flexible for us to model the relationship between different features.

In relation to the singularity issue, kernel matrix also has its advantages. When $d\geq n$ is true, where $d$ is the number of the features and $n$ is the dimension of them, some feature representation like covariance matrix is bound to be singular. In contrast, kernel matrix can handle this situation well. A direct application of Micchelli’s Theorem \cite{KM2} gives the following result for this case.

\textbf{Theorem 1}. \emph{Let $f_1$, $f_2$, $\cdot \cdot \cdot$, $f_d$ be a set of different n-dimensional vectors. The matrix $M_{d\times d}$ computed with a RBF kernel $\kappa (f_i , f_j)=exp(-\gamma ||f_i-f_j||^2)$ is guaranteed to be nonsingular, no matter what values d and n are.}

According to Micchelli’s Theorem, the RBF kernel also satisfies the above theorem to ensure the nonsingularity of D-K features. The presence of the kernel matrix as feature representation provides us great freedom to choose the most appropriate one for a kernel representation.
%Lastly, in case we cannot be sure about the nonsingularity for a kernel matrix, we can always analyze it with the definition of positive definiteness and/or append a regularizer to this matrix as a preemptive measure.

\subsubsection{Kernel Function}
To reduce the computational complexity, we use the commonly used RBF kernel Eq.(10) as the kernel of our matrix for its superior properties. Given $n$ $d$-dimensional vectors, $x_1, \cdot \cdot \cdot , x_n$, computing all the entries $||f_i-f_j||^2 (i, j = 1, \cdot \cdot \cdot , d)$ has the complexity of $O(nd^2)$. In addition, the proposed kernel representation based on RBF kernel could be quickly computed via integral images. Noting that $||f_i-f_j||^2=f_i^\top f_i-2f_i^\top f_j+f_j^\top f_j$, we can precompute $d^2$ integral images for the inner product of any two feature dimensions.

Generally, the availability of more samples makes kernel evaluation more reliable. In RBF kernel function, more samples make the parameters converge towards their true values. However, in practice we are constrained by the number of available training samples. Also, the proposed kernel matrix has a fixed size $(d\times d)$, independent of the number of samples $(n)$ in a set. Due to this, the kernel-based representations obtained from two different-sized sets can be directly compared.

\subsection{Diffusion-kernel (D-K) Features}
The ADKMM includes two processes. First, we input the face video clip $C_o$ or images $I_o$, the anisotropic diffusion method diffuse the input $C_o$ or $I_o$ to enhance edge information. After several diffusion iterations, the edge of the face images will be preserved and become sharper. From the diffused video clip $C_{dif}$ or images $I_{dif}$, we can obtain more depth information and boundary locations of face images.

Next, we extract D-K features from these diffused face video clip $C_{dif}$. As we use the RBF kernel Eq.(10) as the kernel function, our model is defined as
$$f=\kappa (C_{dif},C_{dif})=exp(-\gamma ||C_{dif}-C_{dif}||^2). \eqno{(11)}$$
We vectorize pixel values of each frame of $C_{dif}$ as a column vector, and the video clip $C_{dif}$ is represented as a matrix $M_{d\times n}$, where $d$ is the dimension of the images and $n$ is the number of the frames. Since the dimension $d$ of the matrix is so high that will cost huge computing resource and time, we then reduce the dimension as $lowd$. After the dimensionality reduction of every frame, the matrix $M_{d\times n}$ which represent the diffused face video clip $C_{dif}$ becomes $M_{lowd\times n}$. Then we input the low dimension matrix as the representation of $C_{dif}$ into the model and gain a $M_{lowd\times lowd}$ D-K feature. The D-K features can guide to distinguish fake face images from real one effectively since they reflect the inner correlation of sequential face images like $C_{dif}$.

\subsection{Deep Features}
Deep learning algorithms have been successfully applied in several vision tasks such as face detection \cite{CNN1} and face recognition \cite{CNN2}. CNNs are designed to extract the local features by combining three architectural concepts that perform some degree of shift, scale, distortion invariance, local receptive fields, shared weight and subsampling. The ability of both convolution layers and subsampling layers to learn distinctive features from the diffused image helps to extract features and achieve better performance for face liveness detection.

%\subsubsection{Convolution Neural Network Feature}
The CNN is pre-trained on the several datasets which totally has more than 500 thousand images of 80 clients to obtain good initializations. The pre-trained model we used  is the AlexNet \cite{Alexnet} for its impressive performance. The AlexNet contains convolutional layers, normalization layers, linear layers, ReLU activation layers, and max-pooling layers. For simplicity, we use L1-5 to represent the 5 convolutional layers, and L6-8 describe the 3 linear layers. The L3-5 are connected to one another without any intervening pooling or normalization layers. The fully-connected layers L6-8 have 4096 neurons each. The L6-7 output features with the dimension of 4096, and the dimensionality of features in L8 is 1000. The L8 is followed by a softmatx classifier to generate probability distribution for classification. Previous studies \cite{Alexnet, L7} show that the 4096-dimensional features of L7 perform better than many handcrafted features. In our network, the L1-7 layers are used as the feature extractor, and we use the 4096-dimensional features of L7 as the deep features.

As mentioned above, whether the input is a face image $I_o$ or a video clip $C_o$, the ADKMM can diffuse and extract the D-K features from them. When the input is a video clip$C_o$, we randomly select a frame to represent the whole video clip. We assume that the deep feature of this image frame can represent the deep feature of input video clip $C_o$.

\subsection{Generalized Multiple Kernel Learning}
Multiple kernel learning refers to a set of machine learning methods that use a predefined set of kernels and learn an optimal linear or non-linear combination of kernels as part of the algorithm. Multiple kernel learning method has the ability to select for an optimal kernel and parameters from a larger set of kernels, reducing bias due to kernel selection while allowing for more automated machine learning methods. Instead of creating a new kernel, multiple kernel algorithms can be used to combine kernels already established for each individual data source.
%\subsubsection{Classification}

\begin{figure}[htb]
\begin{minipage}[b]{1.0\linewidth}
  \centering
  \centerline{\includegraphics[width=3.2 in]{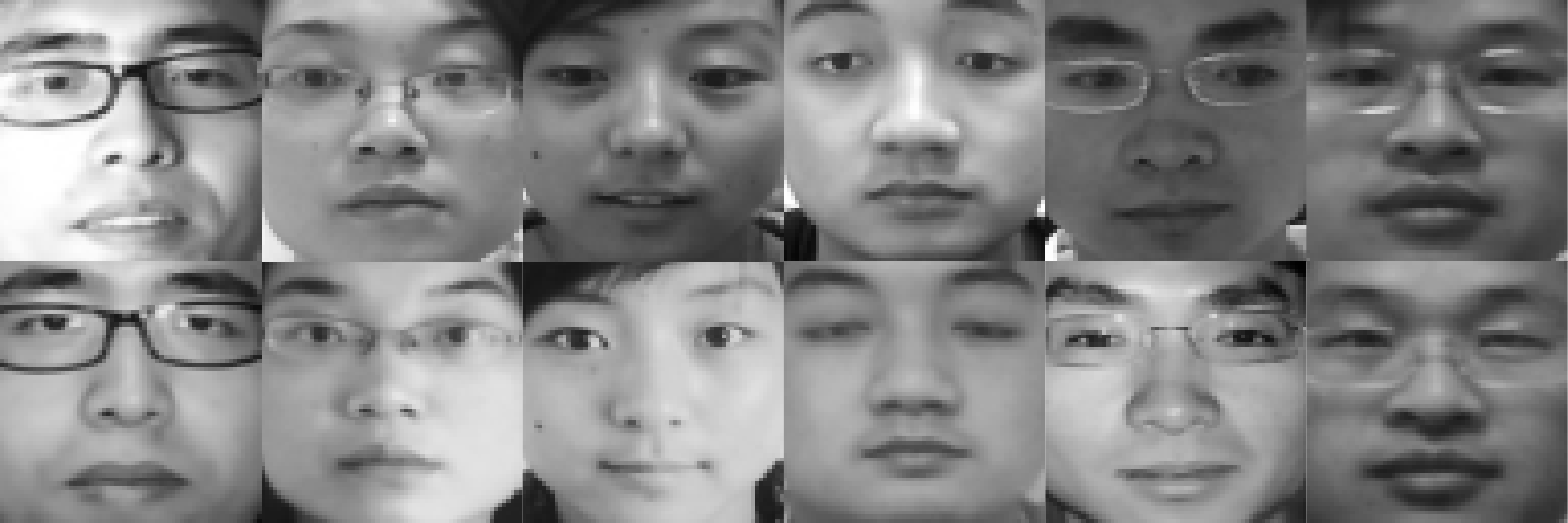}}
\end{minipage}
\caption{Samples from the NUAA dataset (upper row: live faces; lower row: fake faces).}
\end{figure}
\begin{figure}[htb]
\begin{minipage}[b]{1.0\linewidth}
  \centering
  \centerline{\includegraphics[width=3.2 in]{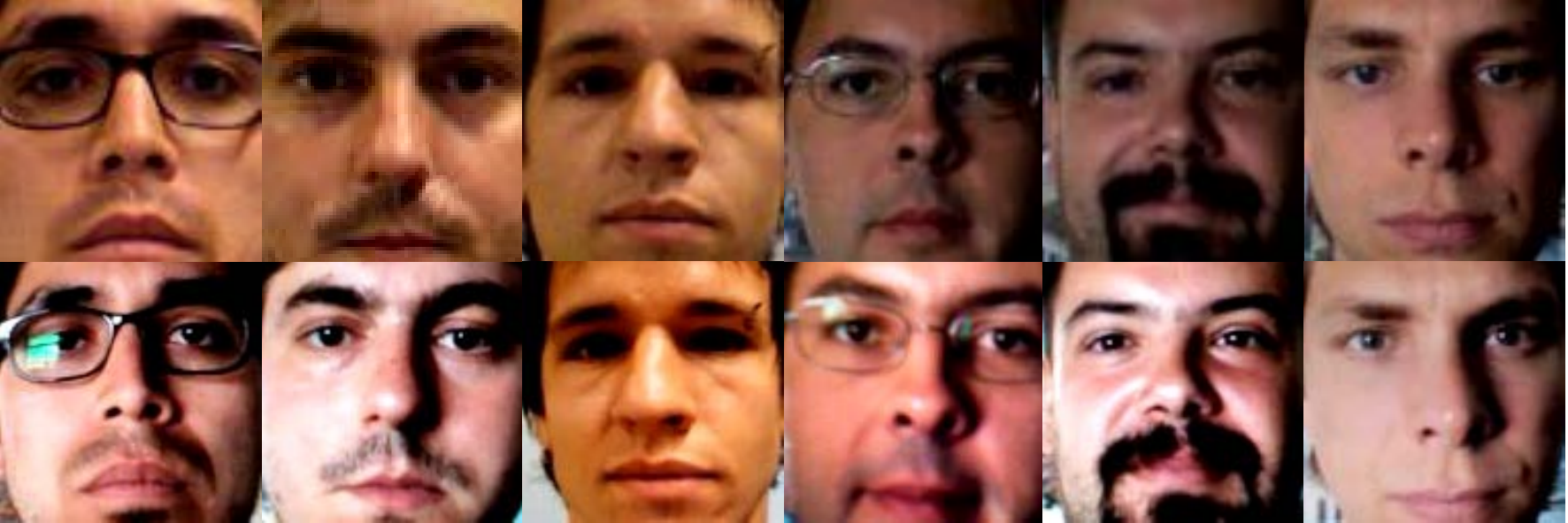}}
\end{minipage}
\caption{Samples from the Replay-Attack dataset (upper row: live faces; lower row: fake faces).}
\end{figure}

Given training one kind of features $\{x_i|i = 1,2,\cdot \cdot \cdot ,N\}$ and the other kind of features $\{y_i|i = 1,2, \cdot \cdot \cdot ,N\}$, the generalized multiple kernel learning is used to fuse these two kinds of training features and train multiple binary classifiers for face liveness detection. The decision function is given by
$$f(x,y) = c_1w^T_1\phi_1(x)+c_2w^T_2\phi(y)+b, \eqno{(12)}$$
where $c_1$ and $c_2$ are the combination coefficients of the two kinds of features with the constraints that $c_1 + c_2 = 1$ and $c_1, c_2 \geq 0$, $w$ and $b$ are parameters of the standard SVM, and $\varphi(\cdot)$ is the a function mapping those two kinds of features to high dimensional space. $c, w$ and $b$ are learned by solving
\begin{equation}
\begin{split}
&\min_{w,b,c}\frac{1}{2}\sum^2_{t=1}\left(c_t||w_t||^2+c^2_t\right)+C\sum^N_il(p_i,f(x_i,y_i))\\
&s.t. \ \ c_1+c_2=1,c_1,c_2\geq0,
\end{split}
\tag{13}
\end{equation}
where $l(p_i, f(x_i, y_i)) = max(0,1-p_if(x_i, y_i))$ is the loss function, and $p_i = \{+1,-1\}$ is the label of the $i$-th training sample. Similar to \cite{GMKL}, Eq.(12) can be reformulated by replacing the SVM with its dual form:
\begin{equation}
\begin{split}
&\min _{c} \frac{1}{2}(c^2_1+c^2_2)+J(c_1,c_2)\\
&s.t. \ \ c_1+c_2=1,c_1,c_2\geq0,
\end{split}
\tag{14}
\end{equation}
where
\begin{equation}
\begin{split}
&J(c_1,c_2)=\\
&\max_{\alpha}\sum^N_{i=1}\alpha_1-\frac{1}{2}\sum^N_{i,j=1}\alpha_i\alpha_j p_i p_j(c_1k_1(x_i,x_j)+c_2k_2(y_i,y_j))\\
&s.t. \ \ \sum^N_{i_1}\alpha_ip_i=0,0\leq \alpha_i \leq C, i = 1,2,\cdot \cdot \cdot , N,
\end{split}
\tag{15}
\end{equation}
$\alpha$ is the dual variable, $k_1(\cdot , \cdot)$ and $k_2(\cdot , \cdot)$ are kernel functions for two kinds of training features, respectively. Here, the RBF kernel function $k_1(x_i,x_j)=exp(-\gamma||x_i-x_j||^2)$and the linear kernel function $k_2(y_i, y_j) = y^T_i y_j$ are used, where $\gamma > 0$ is the kernel parameter. Following \cite{GMKL}, Eq.(13) is solved by iteratively updating the linear combination coefficients $c$ and the dual variable $\alpha$.

\section{Experimental Results}
In this section, we perform an extensive experimental evaluation on various datasets to validate the effectiveness and the superiority of our method. We first introduce two benchmark datasets: the NUAA dataset and the Replay-Attack dataset. Then, we give the detailed description of parameter choices for the ADKMM and generalized multiple kernel learning method. Finally, we compare our method with a number of face liveness detection methods and demonstrate the outstanding performance of the proposed method.

\subsection{Datasets}
\textbf{NUAA}: The NUAA dataset \cite{sta3} is publicly available and the most widely adopted benchmark for the evaluation of face liveness detection. The database images consist of 15 different clients which provide 12,614 images of both live and photographed faces. These images were resized to $64\times 64$ pixels with gray-scale representation. Some samples of the NUAA dataset are shown in Figure 4. For the training set, a total of 3,491 images (live:1,743 / fake:1,748) were selected, while the testing set was composed of 9,123 images (live:3,362 / fake:5,761).

 \textbf{Replay-Attack}: The Replay-Attack dataset \cite{RAdataset} was released in 2012 and is publicly available and widely used. It consists of 1,300 video clips of 50 different subjects. These video clips are divided into 300 real-access videos and 1000 spoofing attack videos which the resolution is $320\times 240$ pixels. The dataset takes into consideration the different lighting conditions used in spoofing attacks. Some samples of the Replay-Attack dataset are shown in Figure 5. Note that the Replay-Attack database is divided into three subsets: training, development and testing.

\begin{table}[tbp]
\centering  % 表居中

\begin{tabular}{cc}  % {lccc} 表示各列元素对齐方式，left-l,right-r,center-c
\hline
Iteration numbers $(L)$ &Accuracy \\
\hline  % \hline 在此行下面画一横线
5 &94.2\% \\         % \\ 表示重新开始一行
10 &98.7\% \\        % & 表示列的分隔线
15 &\textbf{99.3\% }\\
20 &96.5\% \\
25 &93.1\% \\
30 &90.6\% \\
\hline
\end{tabular}
\caption{Performance with different iteration numbers $(L)$ on the NUAA dataset.}
\end{table}

\begin{table}[tbp]
\centering  % 表居中

\begin{tabular}{lc}  % {lccc} 表示各列元素对齐方式，left-l,right-r,center-c
\hline
Methods &Accuracy \\
\hline  % \hline 在此行下面画一横线
Ours &99.3\% \\
ND-CNN \cite{diffCNN} &99.0\% \\
DS-LSP \cite{diffspeed}&98.5\% \\
CDD \cite{CDD}&97.7\% \\
DoG-SL \cite{sta4}&94.5\% \\
M-LBP \cite{sta5}&92.7\% \\
DoG-LRBLR \cite{sta3}&87.5\% \\
DoG-F \cite{sta1}&84.5\% \\         % \\ 表示重新开始一行
DoG-M \cite{sta2}&81.8\% \\        % & 表示列的分隔线
\hline
\end{tabular}
\caption{Performance comparison on the NUAA dataset with other methods.}
\end{table}

\subsection{Parameter Settings}
All parameters of our method are found experimentally and remain unchanged for all datasets. In our anisotropic diffusion scheme, we use $c(\cdot)=exp(-\bigtriangledown / K^2)$ as the the gradient value function and the constant $K$ was fixed as 15. We fixed the $\lambda = 0.15$ in Eq.(6). In our generalized multiple kernel learning method, the combination coefficients $c_1,c_2$ are initialized as $1/2$. Since the face liveness detection is a binary classification problem, parameter $\lambda$ in the RBF kernel function $k_1(x_i,x_j)$ is fixed as $1/2$.
\subsection{Performance Evaluation on the NUAA Dataset}
We evaluate the performance of our approach on the NUAA dataset and conducted many experiments with different iteration numbers $(L)$ in Eq.(6) as shown in Table 1. The best detection accuracy achieved using the NUAA dataset was 99.3\% using value of $L=15$. From Table 1, we can see that increase the number of iterations does not always lead to higher accuracy. For example, experiments where $L=10$ resulted in the accuracy of 98.7\%, and experiments where $L=25$ resulted in an accuracy of 93.1\%.

To prove the effectiveness and superiority of the proposed method, we compared the performance of our approach with all previously proposed approaches on the NUAA dataset.

The compared approaches include multiple difference of Gaussian (DoG-M) \cite{sta2}, DoG and high frequencybased (DoG-F) \cite{sta1}, DoG-sparse low-rank bilinear logistic regression (DoG-LRBLR) \cite{sta3}, multiple local binary pattern (M-LBP) \cite{sta5}, DoG-sparse logistic (DoG-SL) \cite{sta4}, component-dependent descriptor (CDD) \cite{CDD}, diffused speed-local speed pattern (DS-LSP) \cite{diffspeed} and nonlinear diffusion based convolution neural network (ND-CNN) \cite{diffCNN}. Owing to the strong performance of the ADKMM, our method can model the differences between live and fake faces efficiently. As shown in Table 2, our method achieves the best performance with an accuracy of 99.3\% beyond the previous approaches.
\subsection{Performance Evaluation on the Replay-Attack Dataset}
We describe our performance evaluation on the Replay-Attack dataset, which is designed specifically for face spoofing studies and contains diverse spoofing attacks as well. We employed $L=15$ for our ADKMM achieving the best performance on the NUAA dataset. Besides training and testing samples, the Replay-Attack dataset also provides development samples to efficiently evaluate the performance of anti-spoofing methods. For accurately measure the performance on the Replay-Attack dataset, we computed the half total error rate (HTER) \cite{HTER} to measure the performance of our proposed approach. The HTER is half of the sum of the false rejection rate (FRR) and false acceptance rate (FAR):
$$HTER=\frac{FRR+FAR}{2}$$

A performance comparison with previously proposed methods is shown in Table 3. On ReplayAttack-test set, the HTER of our method is 4.30\% and our HTER is 5.16\% on ReplayAttack-devel set. From table 3, we can know that the HTER of our method is better than that of LBP-based methods \cite{RAdataset, sta5}. It indicates that the ADKMM can evidently estimate the difference in surface properties between live and fake face images. The result of our method is also better than other diffusion-based methods \cite{diffspeed, diffCNN} which indicates the D-K features also have the ability to reflect the difference between face image sequences in inner correlation. The detail results shown in Table 3 confirm that the proposed method with ADKMM achieves an impressive accuracy under various types of spoofing attacks as compared to previous approaches.

\begin{table}[tbp]
\centering  % 表居中

\begin{tabular}{l|c|c}  % {lccc} 表示各列元素对齐方式，left-l,right-r,center-c
\hline
Methods &devel&test\\
\hline  % \hline 在此行下面画一横线
LBP$^{u2}_{3\times3}+x^2$ \cite{RAdataset}&31.24\%&34.01\% \\
LBP$^{u2}_{3\times3}$+LDA \cite{RAdataset}&19.60\%&17.17\% \\
LBP$^{u2}_{3\times3}$+SVM \cite{RAdataset}&14.84\%&15.16\% \\
LBP+SVM \cite{sta5}&13.90\%&13.87\%\\
DS-LBP \cite{diffspeed}&13.73\%&12.50\%\\
ND-CNN \cite{diffCNN}&--&10\%\\
Ours&5.16\%&4.30\%\\
\hline
\end{tabular}
\caption{Performance comparison using HTER measure on the Replay-Attack dataset.}
\end{table}

\section{Conclusions}
In this paper, we have presented an anisotropic diffusion-based kernel model (ADKMM) for face liveness detection. The anisotropic diffusion method helps to enhance edge information and boundary locations of face images. Diffusion-kernel (D-K) features extracted from these images can significantly represent the differences in surface properties and inner correlation between live and fake face images. The D-K features and the deep features are fused by a generalized multiple kernel learning method, thus achieves excellent performance against face spoofing attack. The ADKMM allows us to detect face liveness even under different lighting conditions usually used in attack attempts. Experimental results compared with other face liveness methods show the superiority and outstanding performance of our method.

{\small
\bibliographystyle{ieee}
\bibliography{egbib}
}

\end{document}